\documentclass{article}
\usepackage{ijcai25}

\usepackage{times}
\usepackage{soul}
\usepackage{url}
\usepackage[utf8]{inputenc}
\usepackage[small]{caption}
\usepackage{graphicx}
\usepackage{amsmath}
\usepackage{amsthm}
\usepackage{booktabs}
\usepackage[ruled,vlined]{algorithm2e}
\SetKwInput{KwInput}{Input}
\usepackage[switch]{lineno}
\usepackage{bm}
\usepackage{amsmath}
\usepackage{enumitem}
\usepackage{amssymb}
\usepackage{setspace}
\usepackage{multirow}
\usepackage{pifont}
\usepackage{tikz}
\usepackage{xcolor}
\usepackage{subcaption}
\usepackage{tikz}
\definecolor{tan}{rgb}{0.937, 0.902, 0.843}
\usepackage{colortbl} 
\usepackage{pifont}
\usepackage{amsmath,amssymb,bm}

\title{FCKT: Fine-Grained Cross-Task Knowledge Transfer with Semantic Contrastive Learning  for Targeted Sentiment Analysis}

\author{
	Wei Chen$^1$
	\and
	Zhao Zhang$^2$\and
	Meng Yuan$^{1}$\and
	Kepeng Xu$^3$
	\And
	Fuzhen Zhuang$^{1,4\dagger}$\\
	\affiliations
	$^1$School of Artificial Intelligence, Beihang University,  China\\
	$^2$School of Computer Science and Engineering, Beihang University, China\\
	$^3$Xidian University, China \quad
	$^4$Zhongguancun Laboratory,  China\\
	\emails
	\{chenwei23, zhuangfuzhen\}@buaa.edu.cn
}

\begin{document}

\maketitle
\let\oldthefootnote\thefootnote
\renewcommand{\thefootnote}{\relax} 
\footnotetext{$^\dagger$ Corresponding author.} 
\let\thefootnote\oldthefootnote

\begin{abstract}
	In this paper, we address the task of targeted sentiment analysis (TSA), which involves two sub-tasks, i.e., identifying specific aspects from reviews and determining their corresponding sentiments. Aspect extraction forms the foundation for sentiment prediction, highlighting the critical dependency between these two tasks for effective cross-task knowledge transfer.
	While most existing studies adopt a multi-task learning paradigm to align task-specific features in the latent space, they predominantly rely on coarse-grained knowledge transfer. Such approaches lack fine-grained control over aspect-sentiment relationships, often assuming uniform sentiment polarity within related aspects. This oversimplification neglects contextual cues that differentiate sentiments, leading to negative transfer.
	To overcome these limitations, we propose FCKT, a fine-grained cross-task knowledge transfer framework tailored for TSA. By explicitly incorporating aspect-level information into sentiment prediction, FCKT achieves fine-grained knowledge transfer, effectively mitigating negative transfer and enhancing task performance.
	Experiments on three datasets, including comparisons with various baselines and large language models (LLMs), demonstrate the effectiveness of FCKT. The source code is available on https://github.com/cwei01/FCKT.
	
\end{abstract}

\section{Introduction}
\label{introduction}
Unlike traditional sentiment analysis tasks~\cite{Bing2012Sentiment}, aspect-based sentiment analysis (ABSA)~\cite{wang2016recursive,tang2020dependency} requires a deeper contextual understanding and the extraction of more fine-grained information. ABSA has been widely applied in tasks such as customer feedback analysis and academic paper review evaluation~\cite{nath2024aspect}.
ABSA typically involves identifying (1) aspects (entities or attributes discussed), (2) opinions (sentiment expressions), and (3) sentiment polarity. However, research indicates that over 30\% of sentiment expressions are implicit~\cite{cai2021aspect,chen2024modeling}, meaning the ABSA must infer aspects and sentiments from the context without  predefined opinions. 
In this paper, we address these challenges by developing targeted sentiment analysis (TSA).
It involves two sub-tasks, i.e., identifying relevant aspects from reviews and explicitly associating them to their corresponding sentiments~\cite{chen2022hierarchical,zhou2024comprehensive}.

\begin{figure}[t]
	\centering
	\setlength{\fboxrule}{0.pt}
	\setlength{\fboxsep}{0.pt}
	\fbox{
		\includegraphics[width=0.98\linewidth]{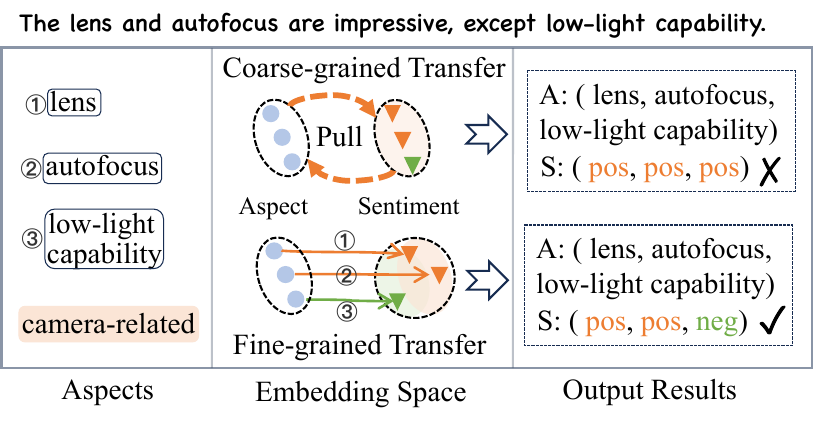}
	}
	\caption{
		Illustration comparing coarse-grained and fine-grained transfer methods. Coarse-grained transfer assumes uniform sentiment polarity for related aspects, while fine-grained transfer captures nuanced aspect-sentiment relationships for precise alignment.}
	\label{intro}
\end{figure}

When TSA was first proposed, a two-stage pipeline method was developed~\cite{hu2019open,kalbhor2023survey}. However, this approach often lead to error propagation between sub-tasks. Moreover, aspects are intrinsically linked with sentiment, and this staged pipeline approach disrupts this crucial interaction~\cite{xu2021learning}.
To address these issues, many end-to-end approaches have been proposed~\cite{chen2022enhanced,li2023dual,zhu2024pinpointing}, aiming to jointly extract aspect terms and classify their sentiments by modeling the interactions between aspects and sentiments.
A widely studied technique in this field is the task-specific feature alignment approach~\cite{chen2024modeling}.
This method involves two key steps: first, encoding task-specific features for both aspects and sentiments; second, aligning these features in the latent space, even when some aspects differ in sentiments, to enhance cross-task knowledge interaction.

Despite recent advancements, these methods still rely on coarse-grained knowledge transfer, which fails to provide explicit and fine-grained control over aspect-sentiment relationships. This limitation often leads to negative transfer, as coarse-grained approaches assume uniform sentiment polarity across related aspects,  ignoring key contextual differences. For instance, as illustrated in Figure~\ref{intro}, coarse-grained methods treat all camera-related aspects (``lens", ``autofocus",  and ``low-light capability") share the same sentiment representation. Consequently, these methods incorrectly classify ``low-light capability" as positive, overlooking the negative sentiment implied by the contrastive cue ``except". This misalignment, stemming from the lack of explicit control over knowledge transfer, is particularly problematic when aspect-sentiment relationships are subtle or context-dependent.

Along this research line, explicit fine-grained knowledge transfer has emerged as a promising solution to enhance the interaction between aspects and sentiments. It disentangles aspect-sentiment relationships and enables dynamic alignment tailored to each aspect. By isolating sentiment representations for each aspect, it accurately associates “lens" and “autofocus" with positive sentiment and “low-light capability" with negative sentiment, leveraging contextual cues like the contrast implied by “except". This ensures more precise and context-aware sentiment classification.
However, despite its potential, explicit fine-grained cross-task knowledge transfer faces the more severe challenge of error accumulation that must be addressed:
(1)
In fine-grained knowledge transfer, the performance of tasks such as aspect extraction and sentiment classification is highly interdependent, as information flows directly between tasks. While leveraging aspect extraction to assist sentiment classification fosters cross-task collaboration, it also introduces the risk of error propagation. Errors or ambiguities in aspect extraction, such as incorrectly identified or missing aspects, are directly transferred to the sentiment classification phase, compounding inaccuracies in the final predictions. To address this, a more effective approach is required to constrain aspect extraction, ensuring more accurate identification and minimizing the influence of ambiguity.
(2)
Moreover, even with perfect aspect extraction, fine-grained knowledge transfer may still struggle due to insufficient supervisory signals for sentiment classification. As the sentiment classifier relies heavily on aspect extraction, its training can become overly dependent on these features, potentially overlooking other crucial contextual cues that are not explicitly tied to aspect information. This lack of comprehensive supervision can hinder the classifier’s ability to generalize effectively. This underscores the need for balanced, holistic supervision to optimize fine-grained knowledge transfer.

To address these challenges, we propose FCKT, a fine-grained cross-task knowledge transfer framework tailored for the TSA task. Specifically, we design a token-level semantic contrastive learning mechanism. In this approach, the start and end tokens of the same aspect are treated as positive pairs, while tokens from unrelated aspects serve as negative pairs. This refines the model's understanding of aspects and enhances its ability to capture subtle contextual dependencies.
To mitigate the lack of supervisory signals in the sentiment classifier, we introduce an alternating learning strategy. This approach trains a fixed proportion of samples using real labels, while the remaining samples are updated based on predictions from the previous model.
This enables FCKT to leverage both real labeled information and predicted transfer knowledge, synergistically improving the effectiveness of TSA.
Furthermore, while Large Language Models (LLMs) like GPT-3.5 and GPT-4 excel in many general-purpose scenarios, they often struggle to adapt to task-specific challenges in fine-grained TSA tasks. Our proposed framework surpasses these models in both few-shot learning and chain-of-thought reasoning scenarios by leveraging task-specific knowledge transfer mechanisms and tailored training strategies. These results highlight the capability of FCKT to address the unique complexities of TSA tasks, providing valuable insights for advancing task-specific methods in the LLM era.
Our key contributions are summarized as follows:
\begin{itemize}[leftmargin=*,labelindent=0pt,topsep=2pt,itemsep=1pt] 
	\item We propose a token-level semantic contrastive learning mechanism that enhances the model's understanding of aspects by treating start and end tokens as positive pairs and unrelated tokens as negative pairs, enabling it to effectively capture subtle contextual dependencies. 
	\item To address insufficient supervisory signals in sentiment classification, we introduce an alternating learning strategy integrates real labeled data with predicted transfer knowledge, significantly improving task-specific performance. 
	\item We conduct extensive experiments on three real-world datasets, demonstrating the superiority of FCKT over existing approaches. We also demonstrate the effectiveness of our approach across various LLM models, showcasing its superior performance in the era of LLMs.
\end{itemize}

\section{Related Work}
\textbf{Targeted Sentiment Analysis.}
Targeted sentiment analysis (TSA) involves extracting aspect terms and their associated sentiment polarities, a key focus in recent research. Early work approached aspect extraction and sentiment analysis as separate tasks, employing methods such as  Long Short-Term Memory (LSTM) networks \cite{luo2019doer} and Bidirectional Encoder Representations from Transformers (BERT)~\cite{yang2020constituency}.
Recent studies have shifted toward end-to-end solutions using multi-task learning frameworks~\cite{luo2019doer,lin2020shared,chen2022hierarchical}, highlighting the interdependence between aspects and sentiments. Techniques such as shared-private feature interaction~\cite{lin2020shared} and task-regularization interaction~\cite{chen2022hierarchical} have been introduced to improve performance by leveraging cross-task knowledge transfer.
However, current approaches mainly rely on coarse-grained alignment approches , which lacks fine-grained control over aspect-sentiment relationships. This often leads to negative transfer, as coarse-grained approaches assume uniform sentiment polarity within related aspects, overlooking contextual cues that differentiate sentiments.
More recently, large language models (LLMs) like GPT have demonstrated strong generalization capabilities, outperforming fine-tuned BERT in some low-resource ABSA scenarios~\cite{wang2023chatgpt,zhou2024comprehensive}. Nevertheless, they continue to underperform compared to state-of-the-art task-specific models, underscoring the need for further adaptation to address the unique challenges of resource-constrained TSA tasks.


\noindent
\textbf{Contrastive Learning.} Contrastive learning has recently achieved significant success in various domains~\cite{jaiswal2020survey,zhang2022contrastive,305b2288,zhong2024spike,zhong2024causal,zhong2025ctd,yuan2025hyperbolic,yuan2025hek,yuan2023knowledge}. In ABSA, several studies have explored its integration into model training to improve performance. For example, Liang~\cite{liang2021enhancing} leveraged contrastive learning to distinguish between aspect-invariant and aspect-dependent features, leading to better sentiment classification. Similarly, Xiong~\cite{xiong2022triplet} proposed a triplet contrastive learning network that combines syntactic and semantic information via an aspect-oriented sub-tree and sentence-level contrastive learning. Chen~\cite{chen2024modeling} further utilized contrastive learning to model adaptive task-relatedness between aspect and sentiment co-extraction.
Different from these works, we leverage contrastive learning specifically for aspect extraction to enhance knowledge transfer in TSA.

\section{Methodology}
\begin{figure}[t]
	\centering
	\setlength{\fboxrule}{0.pt}
	\setlength{\fboxsep}{0.pt}
	\fbox{
		\includegraphics[width=0.99\linewidth]{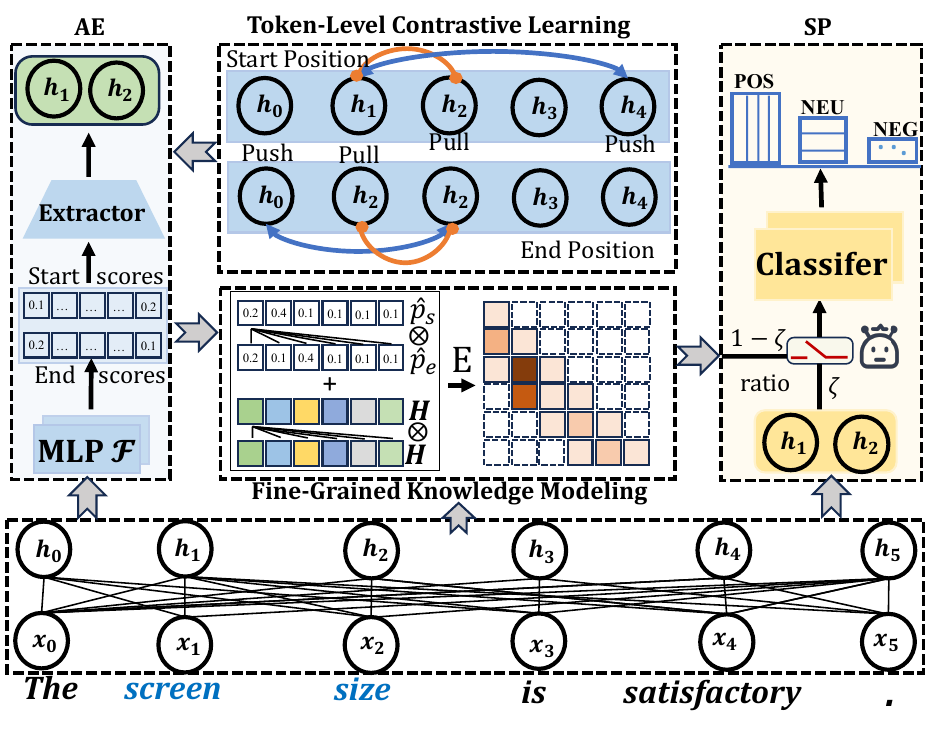}
	}
	\caption{ A depiction of the peopose FCKT framework.
	}
	\label{model}
\end{figure}

In this section, we will introduce proposed FCKT framework in detail, which is depicted in Figure~\ref{model}. 
The comprehensive task consists of two base components: first, extracting opinion aspects (AE), and second, predicting the sentiment polarity (i.e., positive, neutral, or negative)  of each aspect (SP). 

\subsection{Problem Description}
Formally, given a sentence $\bm{x}=\{{x}_{0}, {x}_{1}, \ldots, {x}_{n-1}\}$, where $x_i$ represents the $i^{th}$ token in $\bm{x}$ and $n$ denotes the sentence length, the main objective of FCKT is to extract all possible aspects and predict their corresponding sentiments in $\bm{x}$.

To enhance the representation of contextual semantics, we leverage the widely recognized BERT model~\cite{devlin2019bert}, an effective multi-layer bidirectional Transformer designed to construct rich contextual embeddings by simultaneously capturing both left and right word dependencies. BERT comprises a series of $T$ Transformer layers. The working principle of BERT is summarized as follows: 
\begin{equation} 
	\bm{H}^{0} = W_{i} E + P, \quad \bm{H}^{t} = \mathcal{T} \left(\bm{H}^{t-1}\right) , t \in [1, T], \end{equation} 
where $W_{i}$ is the one-hot representation of the sub-word in the input sentence, $E$ denotes the sub-word embedding matrix, $P$ represents the positional embedding, $\mathcal{T}$ is the Transformer block, and $\bm{H}^{t}$ is the output of the $t^{th}$ Transformer layer.
For the given sentence $\bm{x}=\{{x}_{0}, {x}_{1}, \ldots, {x}_{n-1}\}$, we assume the output of the final Transformer layer is $\bm{H}= \bm{H}^{T}= \{\bm{h}_{0}, \bm{h}_{1}, \ldots, \bm{h}_{n-1}\}\in \mathbb{R}^{n\times d}$, where $d$ is the embedding size.

\begin{figure}[t]
	\centering
	\setlength{\fboxrule}{0.pt}
	\setlength{\fboxsep}{0.pt}
	\fbox{
		\includegraphics[width=0.99\linewidth]{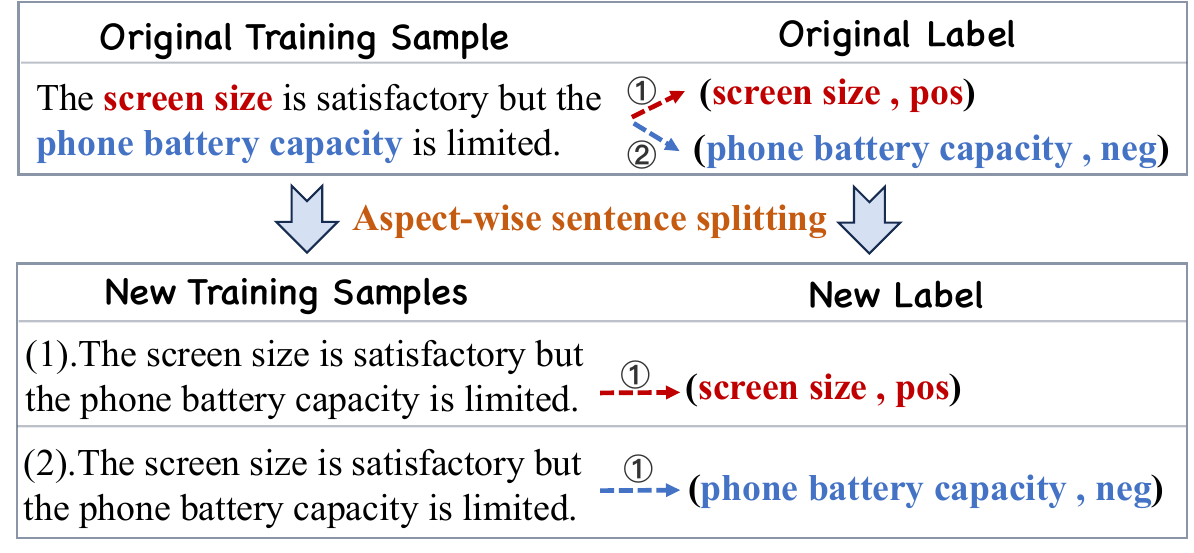}
	}
	\caption{ An example of the sentence splitting process.
	}
	\label{split}
\end{figure}

\textbf{Remark.} In our modeling process, sentences containing multiple aspect terms are split into separate sentences during training, with each focusing on a single aspect, as illustrated in Figure \ref{split}. This preprocessing step allows the model to effectively learn aspect-specific representations by isolating each aspect, avoiding interference from others in the same sentence.
It is important to note that this splitting strategy is only applied during training to facilitate the learning process. During testing, the original sentence structure is preserved, ensuring consistency in evaluation and practical application. Our approach, FCKT, leverages this strategy without altering the original optimization objectives or knowledge transfer mechanisms. \textbf{Further details are provided in the Appendix A}, where we demonstrate that this strategy maintains optimization consistency while enabling fine-grained modeling of aspect-specific interactions.

\subsection{Semantic Contrastive Learning for AE}
In this article, aspects are detected by predicting their start and end boundaries through a linear transformation operation~\cite{lin2020shared}. Specifically, the start and end boundary distributions are predicted as follows: \begin{equation}
\begin{aligned}
	\hat{\bm{p}}_s=\mathcal{F}_{\bm{\phi}_1} \left(\bm{H} \right), 		\hat{\bm{p}}_e=\mathcal{F}_{\bm{\phi}_2} \left(\bm{H} \right),
	\label{s}
\end{aligned}
\end{equation}
where $\mathcal{F}_{\phi_1}$ and $\mathcal{F}_{\phi_2}$ are MLP layers used for extracting aspects. The length of $\hat{\bm{p}}_s$ matches that of the sentence, and each element in $\hat{\bm{p}}_s$ represents the probability of a word being the start of an aspect. It is important to note that this step only predicts the boundary distributions of aspects. To determine the final aspects, a heuristic extraction algorithm~\cite{hu2019open} is subsequently employed.
During optimization, the learning objective for boundary distributions is formulated as: 
\begin{equation}
\begin{aligned}
	\mathcal{L}_{ae} = -\sum_{i=1}^n \{{\bm{p}}_{i,s}^T\log{(\hat{\bm{p}}_{i,s})} + {\bm{p}}_{i,e}^T\log{(\hat{\bm{p}}_{i,e})}\},
	\label{ae}
\end{aligned}
\end{equation}
where $\bm{p}_{i,s}^T\in \mathbb{R}^{n}$ and $\bm{p}_{i,e}^T\in \mathbb{R}^{n}$  are the boundary ground truths (i.e., 0-1 vectors), $\hat{\bm{p}}_{i,s}$ and $\hat{\bm{p}}_{i,e}$ are the predicted boundary distributions, $n$ is the length of sentence.

Compared to conventional coarse-grained transfer methods~\cite{chen2024modeling,sun-etal-2024-minicongts}, fine-grained knowledge transfer imposes stricter requirements on the precision of aspect extraction. Any incorrectly identified or missing aspects are directly transferred to the sentiment classification phase, accumulating prediction inaccuracies.  To mitigate this issue, we introduce a token-level semantic contrastive learning framework to improve aspect representation quality.

Intuitively, tokens within an aspect inherently exhibit strong semantic compatibility, as they frequently co-occur in similar contexts. For instance, as illustrated in Figure~\ref{model}, the end boundary distribution suggests that both the 2nd and 3rd words have comparable probabilities of being the end of an aspect. However, from a semantic perspective, the word “is” is unrelated to the preceding words and is therefore less likely to be part of the aspect boundary. Semantic information can serve as a complementary signal to boundary distributions by verifying whether different tokens form a “semantically reasonable" aspect.
To leverage this property, we treat the start and end tokens of the same aspect as positive pairs, encouraging their representations to be closely aligned. For negative pairs, we construct two types: (1) the start token of the aspect paired with end tokens from other, unrelated aspects, and (2) the end token of the aspect paired with start tokens from unrelated aspects. The InfoNCE loss~\cite{chen2024fairgap,chen2024fairdgcl} is then employed to optimize these token-level embeddings. Formally, this loss is defined as:
\begin{equation}
\begin{aligned}
	\mathcal{L}_{cl} = - \!\!\!\!\!\!\sum_{(s,e) \in \mathcal{E}} \!\log \frac{ \exp(\bm{s}(\bm{h}_s, \bm{h}_e) / \tau)}{\sum\limits \exp(\bm{s}(\bm{h}_s, \bm{h}_i) / \tau) \!+\! \exp(\bm{s}(\bm{h}_e, \bm{h}_j) / \tau)},
	\label{cl}
\end{aligned}
\end{equation}
where $(\bm{h}_s, \bm{h}_e)$ are positive pair embeddings, while $(\bm{h}_s, \bm{h}_i)$ and $(\bm{h}_e, \bm{h}_j)$ are negative pairs. $\tau$ is the temperature parameter, $\bm{s}(\cdot, \cdot)$ denotes the cosine similarity function, and $\mathcal{E}$ represents the set of all positive pairs in each sentence.




\subsection{Fine-Grained Knowledge Transfer for SP}
In sentiment prediction task, the sentiment polarity for a given aspect is determined based on the words within its boundaries. Let the start and end boundaries be denoted as $s$ and $e$, respectively. The sentiment is then predicted as: \begin{equation}
\begin{aligned}
	\hat{\bm{y}}{(\bm{\psi})}=\mathcal{C}_{\bm{\theta}}(\bm{H}_{{s\rightarrow e}}),
\end{aligned}
\label{real}
\end{equation} 
where $ \hat{\bm{y}}{(\bm{\psi})}$ represents the predicted class probability distribution, $\mathcal{C}_{\bm{\theta}}$ is a nonlinear classifier that maps input features to the range $[0,1]$, effectively modeling the likelihood of each sentiment class.  $\bm{H}_{s \to e}$ denotes the aspect representation, aggregated from the start boundary $s$ to the end boundary $e$.


As previously mentioned in Section \ref{introduction} , these two sub-tasks exhibit a logical sequential dependence, implying that aspects can provide valuable signals for sentiment prediction. However, sentiment prediction in previous coarse-grained transfer methods relies on ground truth boundaries, which essentially disconnects the sequential connections between these two sub-tasks. Therefore, we employ aspect features to supervise the sentiment prediction process. In this way, aspect extraction can provide sequentially useful supervised signals for sentiment prediction, while sentiment information can influence aspect detection through backpropagation.

To achieve this, we refrain from using the ground truth boundaries as input. Instead, we derive an expectation of word embedding based on the boundary distributions. Consequently, Eq. (\ref{real}) can be redefined as:
\begin{equation}
\begin{aligned}
	\hat{\bm{y}}{(\bm{\ell})}=\mathcal{C}_{\bm{\theta}} \left(\mathbb{E}_{\left(i\sim \hat{\bm{p}}_s,j\sim \hat{\bm{p}}_e\right)}\bm{H}_{i \rightarrow j}\right),
\end{aligned}
\end{equation}
where $\mathbb{E}$ denotes the expectation operator over the indices $i$ and $j$, which are sampled from the predicted start boundary distribution $\hat{\bm{p}}_s$ and the end boundary distribution $\hat{\bm{p}}_e$, respectively. Furthermore,
since $\hat{\bm{p}}_s$ and $\hat{\bm{p}}_e$ are both discrete distributions, we have the following derivation:

\begin{equation}
\begin{aligned}
	\hat{\bm{y}}{(\bm{\ell})}
	&= \mathcal{C}_{\bm{\theta}}\left(\mathbb{E}_{i \sim \hat{\bm{p}}_s} \left[\mathbb{E}_{j \sim \hat{\bm{p}}_e} \left[\sum_{k=i}^j \bm{H}_{k}\right]\right]\right) \\
	&= \mathcal{C}_{\bm{\theta}}\left(\mathbb{E}_{i \sim \hat{\bm{p}}_s} \left[\sum_{j=1}^n \hat{p}_{e,j} \sum_{k=i}^j \bm{H}_{k}\right]\right) \\
	&= \mathcal{C}_{\bm{\theta}}\left(\sum_{i=1}^n\sum_{j=1}^n\hat{{p}}_{s,i} \cdot \hat{{p}}_{e,j}\sum_{k=i}^j\bm{H}_{k}\right),
\end{aligned}
\label{expectation}
\end{equation}
where we assume $n$ is the sentence's length. In this way, these two sub-tasks enable genuine end-to-end learning.

In practice, to compute Eq.~(\ref{expectation}) , one needs to traverse the whole sentence, and the computational complexity is $\mathcal{O}(n^2)$, which is very difficult to calculate. Fortunately,
this complexity can be reduced based on two facts: (\romannumeral1) The end point is not smaller than the start point. (\romannumeral2) The length of an aspect term is usually not large. We then revise 	$\hat{\bm{y}}{(\bm{\psi})}$ to $\hat{\bm{y}}{(\bm{\ell})}$, as follows:
\begin{equation}
\begin{aligned}
	\hat{\bm{y}}{(\bm{\ell})} =
	&\mathcal{C}_{\bm{\theta}}\left( \sum_{i=1}^n\sum_{j=i}^{i+h}\hat{{p}}_{i,s}\cdot \hat{{p}}_{e,j}\sum_{k=i}^j\bm{H}_{k}\right),
\end{aligned}
\label{n-expectation}
\end{equation}
where $h$ is the maximum length of aspect. Thus, the computation complexity is reduced to $\mathcal{O}(nh)$ (i.e., \!$h \ll n $ ).

Actually, in the testing phase, the sentiment is inferred based on the predicted aspects, which is more aligned with our fine-grained transfer modeling. By inputting the distributional boundaries, the sentiment information can be back propagated to supervise the aspect detection process, and enhance its accuracy. However, if we only base on the distributional input, we  may still face challenges due to insufficient supervisory signals for sentiment classifier.
In such a scenario, providing some ground truth data can reduce the error propagation path, and thus alleviate this problem. Specifically, during training, we use an alternate training strategy: a fixed ratio ($\xi$) of samples are trained based on Eq. (\ref{real}), while the remaining samples ($1-\xi$) are optimized using Eq. (\ref{n-expectation}). 
The model parameters are optimized through the following cross-entropy loss formulation \cite{chen2024modeling}:
\begin{equation}
\mathcal{L}_{sp} = 
-\sum_{i=1}^N \sum_{j=1}^K \bm{y}_{i,j} \log \left(\xi  \cdot \hat{\bm{y}}(\bm{\psi})_{i,j} + (1-\xi) \cdot \hat{\bm{y}}(\bm{\ell})_{i,j}\right),
\label{sp}
\end{equation}
where $N$ represents the number of samples, $K$ indicates the number of sentiment types,
$	\hat{\bm{y}}{(\bm{\psi})}$ and $	\hat{\bm{y}}{(\bm{\ell})}$ are predicted sentiment distributions, and ${\bm{y}}$ is the ground truth. This strategy effectively combines real labels with transferred knowledge, leveraging their synergistic integration to enhance FCKT.

\subsection{Model Optimization}
\textbf{Overall Training.} To enhance the performance of FCKT framework, we jointly optimize the aspect extraction loss $\mathcal{L}_{ae}$, sentiment prediction loss $\mathcal{L}_{sp}$, and contrastive loss $\mathcal{L}_{cl}$. The combined objective function is defined as: 
\begin{equation}
\begin{aligned}
	\mathcal{L}=\mathcal{L}_{ae}+ \mathcal{L}_{sp}+ \lambda\mathcal{L}_{cl},
\end{aligned}
\end{equation}
where $\lambda$ is a trade-off parameter for contrastive loss. The entire optimization process is illustrated in Algorithm 1, with all parameters trained in an end-to-end manner.

\noindent \textbf{Complexity Analysis.} We further analyze the time and space complexities of  FCKT. The time complexity is composed of three main components: (1) the aspect extraction module, with a complexity of $\mathcal{O}(n + n^2)$ per training epoch; (2) the sentiment prediction module, with a complexity of $\mathcal{O}(h)$; and (3) the expectation input module, with a complexity of $\mathcal{O}(nh)$. Consequently, the overall time complexity of FCKT is $\mathcal{O}(n^2 + nh)$.
In terms of space complexity, the model incurs costs primarily from word embeddings and the contrastive learning process, resulting in a total space complexity of $\mathcal{O}(nd + n^2)$, where $d$ represents the embedding size.

\begin{algorithm}[t]
\setstretch{0.99}
\caption{Training Framework}
\KwInput{Training dataset $\mathcal{D}$, learning rate $\alpha$, sampling ratio $\xi$, 
	model parameters $\Theta = \{{\bm{\phi}}_1, {\bm{\phi}}_2, {{\bm{\theta}}}\}$}
\While{the convergence criterion is not met}{
	\For{each batch $\mathcal{B}$ in Dataloader($\mathcal{D}$)}{
		Compute $\hat{\bm{p}}_s$ and $\hat{\bm{p}}_e$ using Eq.~(\ref{s}); \\
		Evaluate aspect extraction loss $\mathcal{L}_{ae}$ via Eq.~(\ref{ae}); \\
		Evaluate contrastive loss $\mathcal{L}_{cl}$ via Eq.~(\ref{cl}); \\
		Sample a random value $p \sim \text{Uniform}(0,1)$; \\
		\If{$p > \xi$}{ 
			Construct predicted $\mathbb{E}_{\left(i\sim \hat{\bm{p}}_s,j\sim \hat{\bm{p}}_e\right)}\bm{H}_{i \rightarrow j}$; \\
			Compute the predicted  $\hat{\bm{y}}{(\bm{\ell})}$ via Eq.~(\ref{n-expectation}); \\
		} 
		\Else  {
			Select real aspect embedding $\bm{H}_{{s\rightarrow e}}$; \\
			Compute the real $\hat{\bm{y}}{(\bm{\psi})}$ via Eq.~(\ref{real}); \\
		}
		Evaluate sentiment loss $\mathcal{L}_{sp}$ via Eq.~(\ref{sp}); \\
		Aggregate total loss $\mathcal{L} = \mathcal{L}_{ae} + \mathcal{L}_{\text{sp}} + \lambda \mathcal{L}_{cl}$; \\
		Update parameter $\Theta = \Theta -\alpha \nabla_{\Theta } \mathcal{L}$. \\
	}
}
\label{algorithm}
\end{algorithm}

\section{Experiments}
In this section, we conduct experiments to answer the following five research questions:
\textbf{RQ1:} How does the proposed FCKT perform overall compared to baseline approaches? 
\textbf{RQ2:} How effectively does FCKT perform across different sub-tasks? 
\textbf{RQ3:} How do the various components of FCKT contribute to the final results? 
\textbf{RQ4:} How do different parameter configurations in the proposed method impact its performance? 
\textbf{RQ5:} Does fine-grained knowledge transfer help address the aforementioned challenges effectively?

\begin{table}[t]
\renewcommand{\arraystretch}{1.2}
\resizebox{0.48\textwidth}{!}{
	\begin{tabular}{cccccc}
		\hline
		Dataset & { \#Sentences} & {\#Aspects} & { \#+} & {\#-} & \#0 \\ \hline \hline
		Laptop & 1869 & 2936 & 1326 & 900 & 620 \\
		Restaurant & 3900 & 6603 & 4134 & 1538 & 931 \\
		Tweets & 2350 & 3243 & 703 & 274 & 2266 \\    \hline
\end{tabular}}
\caption{Statistics of three datasets. ``+/-/0''  denote the positive, negative, and neutral sentiment polarities, respectively.}
\label{table1}
\end{table}

\subsection{Experimental Setup}

\textbf{Datasets}.
\label{data}
To fairly assess our proposed FCKT, we conducted our experiments using three public datasets, Laptop, Restaurant, and Tweets.  The data statistics are presented in Table~\ref{table1}. More dataset details are shown in Appendix B.


\noindent
\textbf{Baselines}. 
We compare three categories of models: \textbf{Pipeline-based models}, \textbf{End-to-End models}, \textbf{LLM-based models}.
More baseline details are listed in Appendix C.

\noindent
\textbf{Evaluation}. We adopt three widely used metrics, precision, recall, and F1 score , to evaluate
the FCKT. Please refer to Appendix D for more metric details.

\noindent
\textbf{Implementation Details}.  All the parameter details are comprehensively provided in Appendix E for further reference.

\begin{table*}[t]
\centering
\renewcommand\arraystretch{1}
\resizebox{1\textwidth}{!}{
	\begin{tabular}{c|l|ccc|ccc|ccc}
		\hline 
		\multicolumn{2}{c|}{\multirow{2}{*}{Model}}                & \multicolumn{3}{c|}{Laptop }                                                    & \multicolumn{3}{c|}{Restaurant}                                                      & \multicolumn{3}{c}{Tweets }                                                   \\ \cline{3-11} 
		\multicolumn{2}{c|}{}                                      & \multicolumn{1}{c}{Prec.} & \multicolumn{1}{c}{Rec.} & \multicolumn{1}{c|}{F1} & \multicolumn{1}{c}{Prec.} & \multicolumn{1}{c}{Rec.} & \multicolumn{1}{c|}{F1} & \multicolumn{1}{c}{Prec.} & \multicolumn{1}{c}{Rec.} & \multicolumn{1}{c}{F1} \\ \hline \hline
		\multirow{4}{*}{\begin{tabular}[c]{@{}c@{}}\textbf{Pipeline}\end{tabular}} & CRF-Pipeline$^\dagger$~\cite{mitchell2013open}       & 0.5969                     & 0.4754                   & 0.5293                  & 0.5228                     & 0.5101                    & 0.5164                   & 0.4297                     & 0.2521                    & 0.3173                  \\
		& NN-CRF-Pipeline$^\dagger$~\cite{zhang2015neural}          & 0.5772                     & 0.4932                    & 0.5319                   & 0.6009                     & 0.6193                    & 0.6100                   & 0.4371                     & 0.3712                    & 0.4006                  \\
		& TAG-Pipeline$^\dagger$~\cite{hu2019open}       & 0.6584                     & 0.6719                    & 0.6651                   & 0.7166                     & 0.7645                    & 0.7398                   & 0.5424                     & 0.5437                    & 0.5426                  \\
		& SPAN-Pipeline$^\dagger$~\cite{hu2019open}       & 0.6946                     & 0.6672                    & 0.6806                   & 0.7614                     & 0.7334                    & 0.7492                   & 0.6072                     & 0.5502                    & 0.5769                  \\
		\hline \hline 
		\multirow{9}{*}{\begin{tabular}[c]{@{}c@{}}\textbf{End-to-End} \end{tabular}}    & SPJM$^\dagger$~\cite{zhou2019span}                & 0.6140                     & 0.5820                    & 0.5976                   & 0.7620                     & 0.6820                    & 0.7198                   & 0.5484                     & 0.4844                    & 0.5144                  \\
		& SPAN-Joint$^\dagger$~\cite{hu2019open}          & 0.6741                     & 0.6199                    & 0.6459                   & 0.7232                     & 0.7261                    & 0.7247                   & 0.5703                     & 0.5269                    & 0.5455                  \\  
		&S-AESC$^\dagger$~\cite{lv2021span}                 &  0.6687                     &    0.6492                   &   0.6588                      &    0.7826                   &     0.7050                  &    0.7418                     &    0.5586                   &    0.5374                   &  0.5473                      \\
		&HI-ASA$^\ddagger$~\cite{chen2022hierarchical}                  & 0.6796                   &  0.6625                  & 0.6709                     & 0.7915                 &    0.7621                 &   0.7765                   &     0.5732               &  0.5622              &        0.5676              \\
		&DCS$^\ddagger$~\cite{li2023dual}                  & 0.6812                   &  0.6640                  & 0.6725                     &  0.7835                     &     0.7751                 &   0.7793                     &     0.5862               &  0.5835                 &        0.5848              \\
		&MiniConGTS$^\ddagger$~\cite{sun-etal-2024-minicongts}                  &   0.7206                   &  0.6725                    & 0.6957                         &  0.7926                &     0.7952                 &   0.7939                   &     0.6164                & 0.5728              &       0.5938               \\
		&PDGN$^\ddagger$~\cite{zhu2024pinpointing}                  &   0.7025                   &  0.6812                    & 0.6921                       &  0.8036                    &     0.7985                &   0.8010                     &    0.6235                &  0.5924               &        0.6076              \\
		&AIFI$^\ddagger$~\cite{chen2024modeling}                  &  0.7105                  & 0.6915               & 0.7009                  &  0.7925                   &     0.8034              &   0.7979                      & 0.6342              &  0.5911                &        0.6119           \\
		\hline  			\hline 
		\multirow{8}{*}{\begin{tabular}[c]{@{}c@{}}{\textbf{LLM-based}}\end{tabular}}    
		&GPT-3.5-turbo Zero-Shot$^\ddagger$                 &   0.3462                    &  0.4065                    & 0.3739                       &  0.6221                   &     0.6605              &   0.6407                     &     0.3750                 &  0.2868                  &        0.3250               \\
		&GPT-3.5-turbo Few-Shot$^\ddagger$                 &   0.3389                    &  0.4452                   & 0.3855                       &  0.5847                    &     0.6605              &   0.6203                    &     0.3812                 &  0.2841                  &        0.3256               \\
		&GPT-3.5-turbo CoT$^\ddagger$                 &   0.3430                    &  0.4581                    & 0.3923                      &  0.6624                    &     0.6420              &   0.6520                     &     0.4233                 &  0.2752                  &        0.3335               \\
		&GPT-3.5-turbo CoT + Few-Shot$^\ddagger$                 &   0.3532                   &  0.4581                    & 0.3989                       &  0.6215                    &     0.6790              &   0.6490                     &     0.3752                 &  0.3012                  &        0.3342               \\
		&GPT-4o  Zero-Shot$^\ddagger$                 &   0.3214                    &  0.4065                    & 0.3590                       &  0.6242                    &     0.5741              &   0.5981                     &     0.2326                 &  0.3704                  &        0.2857               \\
		&GPT-4o Few-Shot$^\ddagger$                 &  0.3299                     &  0.4194                    & 0.3693                       &  0.6571                    &     0.5679              &   0.6093                     &     0.2532                 &  0.3631                 &        0.2984               \\
		&GPT-4o CoT$^\ddagger$                 &   0.3371                    &  0.3806                    & 0.3576                       &  0.6643                    &     0.5741              &   0.6159                     &     0.2823                 &  0.3891                 &        0.3272               \\
		&GPT-4o CoT + Few-Shot$^\ddagger$                 &   0.3622                    &  0.4323                    & 0.3941                       &  0.6842                    &     0.6420             &   0.6624                     &     0.4022                 &  0.3842                  &        0.3930               \\
		\hline  			\hline  
		\multirow{1}{*}{\begin{tabular}[c]{@{}c@{}}{\textbf{Ours}} \end{tabular}}   
		& FCKT               & 0.7599                    & 0.6740                   & \cellcolor{tan}\textbf{0.7144}$^*$                     & 0.8449                    & 0.7877                   & \cellcolor{tan}\textbf{0.8153}$^*$                   & 0.6512                    & 0.5962                 & \cellcolor{tan}\textbf{0.6225}$^*$                 \\
		\hline 
\end{tabular}}
\\ \hspace*{\fill} \\
\caption{The overall performance comparison is conducted on three real-world datasets. The “$\dagger$” denotes results directly taken from the original papers, while “$\ddagger$” indicates results reproduced following the methods described in the original papers. Bold values highlight the best performance, and ${*}$ indicates statistical significance with a $p$-value $\leq 0.05$ compared to the best-performing baseline.}
\label{main}
\end{table*}

\subsection{Main Experimental Results (RQ1)}
We conduct experiments on three public datasets and the comparison results between FCKT and the baselines are presented in Table~\ref{main}. This table shows the following:

Among the baselines, we find that our proposed FCKT consistently achieves the best results in all cases on F1 score. 
More precisely, when compared to the state-of-the-art approach AIFI, FCKT demonstrates notable enhancements in F1 score performance across three distinct datasets, with an average improvement of approximately 1.38\%\footnote{Noted TSA remains a challenging task with minimal progress in recent research, making a 1.38\% improvement significant.}.
These findings strongly suggest that the meticulously crafted FCKT exhibits a potential for superior performance. This can be attributed to two primary factors: first, the inclusion of a fine-grained knowledge transfer framework that effectively captures mutual information between sub-tasks, and second, the introduction of a token-level contrastive learning mechanism that facilitates the extraction of more fitting aspects. 
However, the recall of FCKT is lower than several baseline models. 
One possible reason is that our heuristic extraction method filters out semantically irrational aspects based on the cumulative start/end scores, which may inadvertently exclude some false negatives, leading to a lower recall score.

FCKT demonstrates notable advantages in both performance and computational efficiency compared to advanced LLMs. As illustrated in Tables~\ref{main}, even the state-of-the-art LLM, GPT-4, with its vast number of parameters, fails to deliver satisfactory results for TSA, despite leveraging Few-Shot learning and Chain-of-Thought (CoT)~\cite{wei2022chain} enhancement. Moreover, the use of LLMs incurs substantial computational overhead, further underscoring the efficiency of our approach.
This suggests that improving performance in these low-resource domains with LLMs remains a challenging task. Despite advancements in LLMs, some traditional methods for TSA are still valuable, offering insights that complement the limitations of current techniques.

\begin{table}[t]
\centering
\renewcommand\arraystretch{1.1}
\resizebox{0.465\textwidth}{!}{
\normalsize
\begin{tabular}{l|cccc}
	\toprule 	
	Task & Method & Laptop & Restaurant & Tweets \\ \midrule 	\toprule
	\multirow{8}{*}{\textbf{AE}}
	& HI-ASA\textcolor{blue}{ (COLING22)}      & 0.8424  & 0.8511  & 0.7546  \\
	& DCS\textcolor{blue}{ (EMNLP23)}        & 0.8455  & 0.8462  & 0.7543  \\
	& MiniConGTS\textcolor{blue}{ (EMNLP24)}       & 0.8374  & 0.8432  & 0.7514  \\
	& PDGN\textcolor{blue}{ (ACL24)}       & 0.8485  & 0.8546  & 0.7593  \\
	& GPT-4o CoT + Few-Shot      & 0.5120  & 0.6010  & 0.4724  \\
	& AIFI\textcolor{blue}{ (AAAI24)}        & 0.8511  & 0.8631  & 0.7634  \\
	& \textbf{FCKT}     & \textbf{0.8534}  & \textbf{0.8685}  & \textbf{0.7754}  \\ \midrule \midrule
	\multirow{8}{*}{\textbf{SP}} 
	& HI-ASA\textcolor{blue}{ (COLING22)}          & 0.8531  & 0.9257  & 0.8451  \\
	& DCS\textcolor{blue}{ (EMNLP23)}         & 0.8467  & 0.9212  & 0.8429  \\
	& MiniConGTS\textcolor{blue}{ (EMNLP24)}      & 0.8541  & 0.9194      & 0.8349  \\
	& PDGN\textcolor{blue}{ (ACL24)}       & 0.8519  & 0.9224  & 0.8496  \\
	& GPT-4o CoT + Few-Shot      & 0.7045  & 0.7538  & 0.6323  \\
	& AIFI\textcolor{blue}{ (AAAI24)}        & 0.8594  & 0.9305  & 0.8496  \\
	& \textbf{FCKT}     & \textbf{0.8612}  & \textbf{0.9310}  & \textbf{0.8524}  \\ \bottomrule
	\end{tabular}}
	\caption{The performance comparisons with different methods on aspect extraction (F1 score) and sentiment prediction (accuracy).}
	\label{rq2}
\end{table}

\begin{table}[t]
\centering
\label{tab:performance}
\centering
\renewcommand\arraystretch{1.1}
\resizebox{0.48\textwidth}{!}{
\normalsize
\begin{tabular}{l|ccccccc}
	\toprule
	Task               & AKT & TCL & Laptop & Restaurant & Tweets & Avg. Drop \\ \midrule \midrule
	\multirow{4}{*}{\textbf{AE}} & \ding{51}   & \ding{51}   & \textbf{0.8534}  & \textbf{0.8685}   & \textbf{0.7754}  
	& \tikz[baseline=(X.base)]{\node[inner sep=0pt, anchor=base west] (X) { -};} \\
	& \ding{51}   & \ding{55}   & 0.8420           & 0.8612            & 0.7713           
	& \tikz[baseline=(X.base)]{\node[inner sep=0pt, anchor=base west] (X) {\textcolor{purple}{$\blacktriangledown$} 0.91\%};} \\
	& \ding{55}   & \ding{51}   & 0.8451           & 0.8603            & 0.7714           
	& \tikz[baseline=(X.base)]{\node[inner sep=0pt, anchor=base west] (X) {\textcolor{purple}{$\blacktriangledown$} 0.82\%};} \\
	& \ding{55}   & \ding{55}   & 0.8412           & 0.8568            & 0.7630           
	& \tikz[baseline=(X.base)]{\node[inner sep=0pt, anchor=base west] (X) {\textcolor{purple}{$\blacktriangledown$} 1.45\%};} \\ \midrule \midrule
	\multirow{4}{*}{\textbf{SP}} & \ding{51}   & \ding{51}   & \textbf{0.8612}  & \textbf{0.9310}   & \textbf{0.8524}  
	& \tikz[baseline=(X.base)]{\node[inner sep=0pt, anchor=base west] (X) {{} -};} \\
	& \ding{51}   & \ding{55}   & 0.8502           & 0.9214            & 0.8413           
	& \tikz[baseline=(X.base)]{\node[inner sep=0pt, anchor=base west] (X) {\textcolor{purple}{$\blacktriangledown$} 1.20\%};} \\
	& \ding{55}   & \ding{51}   & 0.8492           & 0.9213            & 0.8381           
	& \tikz[baseline=(X.base)]{\node[inner sep=0pt, anchor=base west] (X) {\textcolor{purple}{$\blacktriangledown$} 1.36\%};} \\
	& \ding{55}   & \ding{55}   & 0.8334           & 0.9189            & 0.8293           
	& \tikz[baseline=(X.base)]{\node[inner sep=0pt, anchor=base west] (X) {\textcolor{purple}{$\blacktriangledown$} 2.38\%};} \\ \midrule \midrule
	\multirow{4}{*}{\textbf{TSA}} & \ding{51}   & \ding{51}   & \textbf{0.7144}  & \textbf{0.8153}   & \textbf{0.6225}  
	& \tikz[baseline=(X.base)]{\node[inner sep=0pt, anchor=base west] (X) {{} -};} \\
	& \ding{51}   & \ding{55}   & 0.6942           & 0.8064            & 0.6153           
	& \tikz[baseline=(X.base)]{\node[inner sep=0pt, anchor=base west] (X) {\textcolor{purple}{$\blacktriangledown$} 1.67\%};} \\
	& \ding{55}   & \ding{51}   & 0.6775           & 0.7842            & 0.6072           
	& \tikz[baseline=(X.base)]{\node[inner sep=0pt, anchor=base west] (X) {\textcolor{purple}{$\blacktriangledown$} 3.86\%};} \\
	& \ding{55}   & \ding{55}   & 0.6724           & 0.7792            & 0.5942           
	& \tikz[baseline=(X.base)]{\node[inner sep=0pt, anchor=base west] (X) {\textcolor{purple}{$\blacktriangledown$} 4.93\%};} \\ \bottomrule
	\end{tabular}}
	\caption{The results of different modules on each task. Note that “\ding{51}/\ding{55} AKT” indicates whether the aspect knowledge transfer strategy is used, while “\ding{51}/\ding{55} TCL” denotes the inclusion or exclusion of the token-level contrastive learning mechanism.}
	\label{ablation}
\end{table}

\subsection{Analysis on Both Sub-Tasks (RQ2)}
To verify model's performance in individual tasks, we conduct a comparative analysis tailored for both sub-tasks in Table \ref{rq2}.
In AE, our proposed FCKT is able to achieve the best overall performances in terms of F1 score. Compared with the other two datasets, the improvement of  model on Tweets is more obvious (+2\%). Since the context length of Tweets is usually shorter, our proposed fine-grained transfer methods are more adequate between two sub-tasks, leading to more significant improvements compared to other two datasets.

For SP, FCKT consistently outperforms other approaches across the three datasets. However, its improvement on the Restaurant dataset is less pronounced compared to the other datasets. We hypothesize that this discrepancy may stem from the relatively high sample diversity in the Restaurant dataset, where the limited interaction information might be insufficient to substantially enhance sentiment prediction.

\begin{figure}[t]
\centering
\begin{subfigure}[!]{0.23\textwidth}
\includegraphics[width=\textwidth]{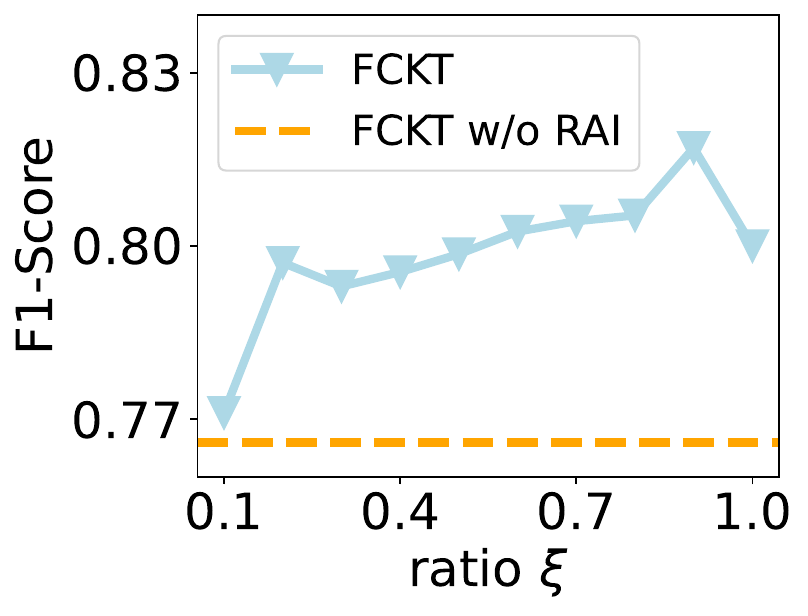}
\end{subfigure} 
\begin{subfigure}[!]{0.23\textwidth}
\includegraphics[width=\textwidth]{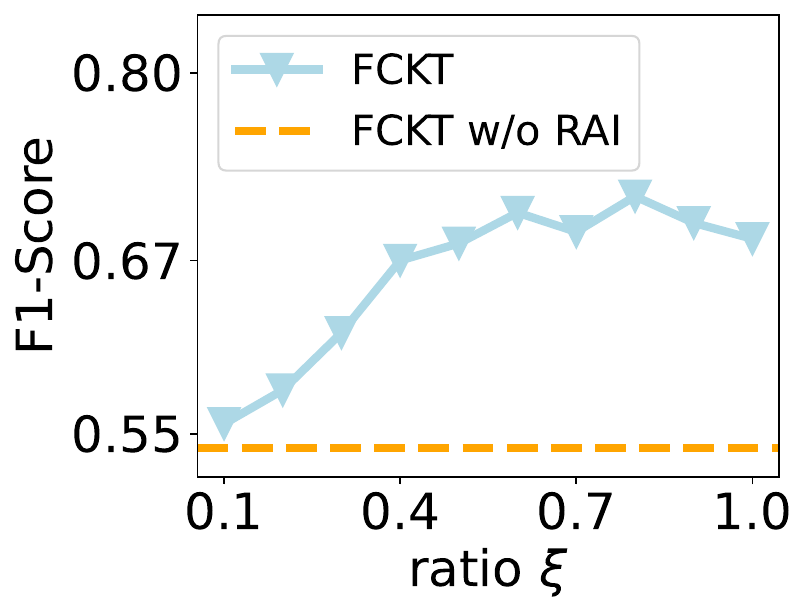}
\end{subfigure}
\caption{The results of FCKT w.r.t different parameter $\xi$ on two datasets. ``FCKT w/o RAI" refers to removing real aspect inputs. }
\label{para1}
\end{figure}

\begin{figure}[t]
\centering
\begin{subfigure}[!]{0.23\textwidth}
\includegraphics[width=\textwidth]{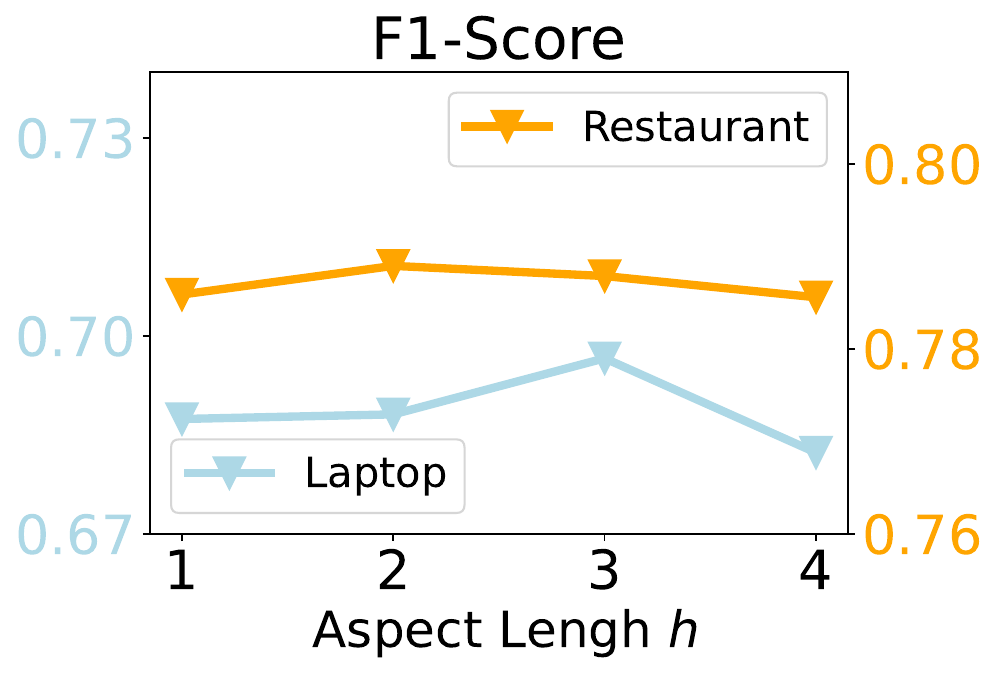}
\end{subfigure} 
\begin{subfigure}[!]{0.23\textwidth}
\includegraphics[width=\textwidth]{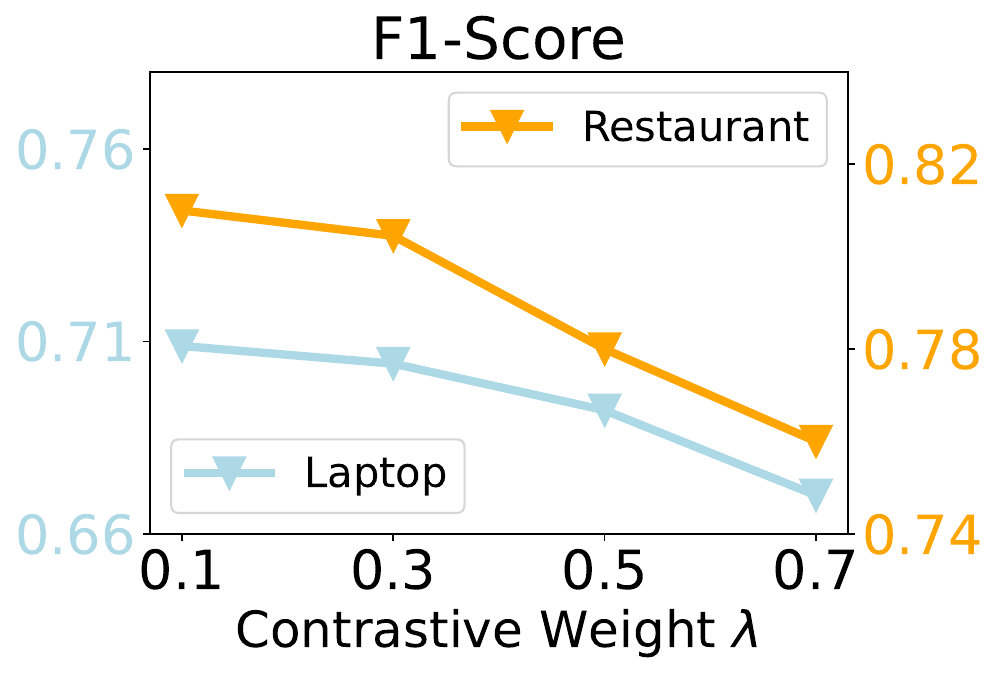}
\end{subfigure}
\caption{The F1 score of FCKT w.r.t varying parameter $h$ and $\lambda$. }
\label{para2}
\end{figure}
\begin{figure}[t]
\centering
\setlength{\fboxrule}{0.pt}
\setlength{\fboxsep}{0.pt}
\fbox{
\includegraphics[width=0.99\linewidth]{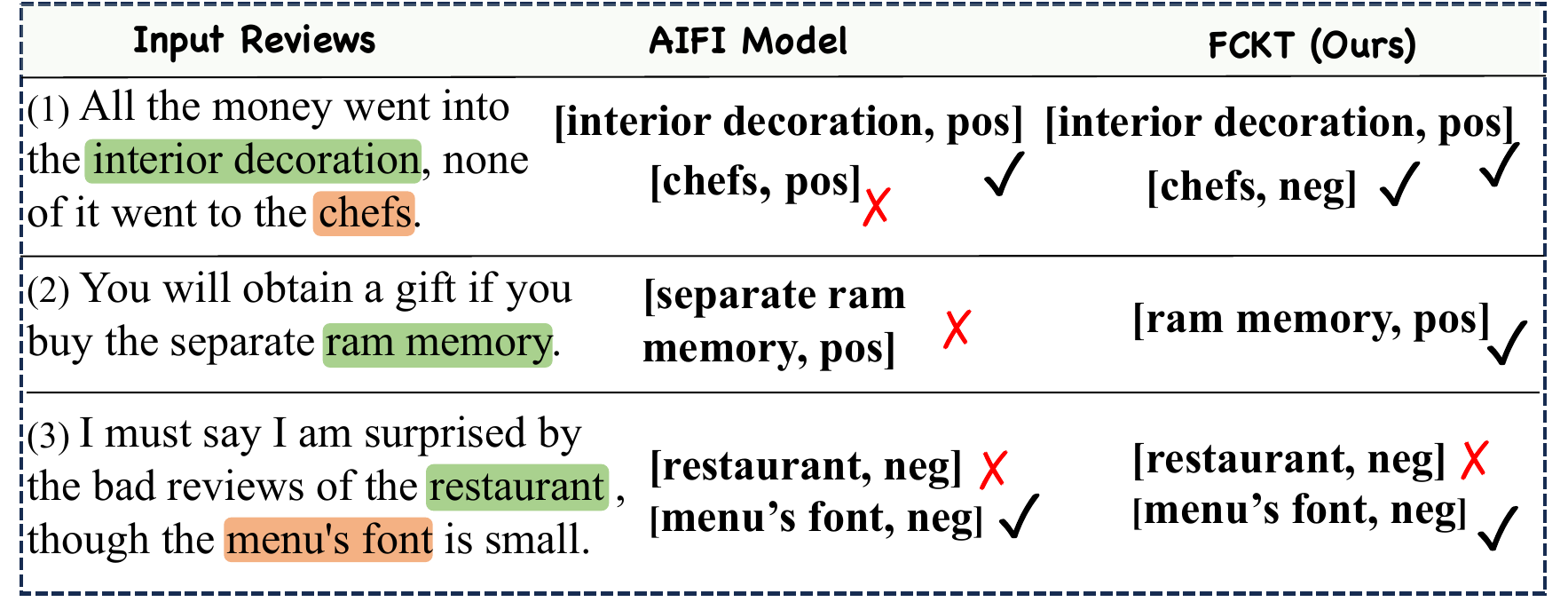}
}
\caption{Some examples of different models. The positive aspects are marked in green, and negative aspects  are marked in orange.  \ding{51} and \ding{55} denote correct and incorrect predictions.
}
\label{case}
\end{figure}
\subsection{Ablation Study  (RQ3)}
To examine the contributions of various components, we delve further into FCKT and carry out ablation studies.
The results are shown in Table~\ref{ablation}. 
It is evident that the removal of specific modules leads to a decrease in model performance, highlighting the indispensable nature of each module. This underscores the critical role that various model components play in achieving optimal performance.
If we delve deeper into the comparisons in TSA task, we can find the “AKT” module holds greater significance, indicated by its lower F1 score compared to the other module. This is not surprising, as cross-task knowledge transfer is the core focus of our work, emphasizing the importance of effectively leveraging mutual information between tasks to enhance performance in TSA.

\subsection{Parameter Analysis (RQ4)}
\textbf{Effect of parameter $\xi$}. 
In FCKT, a critical parameter is the ratio $\xi$, which controls the connection between different tasks. This parameter determines how much of the predicted aspect knowledge is transferred to the second task. We tune $\xi$ within the range of 0 to 1, and the results, based on the TSA task, are presented in Figure~\ref{para1}.
Across these datasets, we observe that the F1 score initially increases rapidly, then stabilizes, and finally declines as $\xi$ increases from 0.8 or 0.9 to 1. 	The reason lies in that when $\xi$ is small, the model lacks sufficient supervision from the ground truth, leading to suboptimal performance. And when $\xi$ is very large, the model fails to leverage the valuable interactions between the two sub-tasks, also resulting in unsatisfactory performance.
We observe that the best performances are usually achieved when $\xi$ is moderate on two datasets,
which implies that neither the ground truth boundary nor distributional boundary is dominantly superior, and a mixture of them can be more favorable.

\noindent
\textbf{Effect of parameter $h$}.
In this study, we investigate the influence of the parameter $h$ as it regulates the length of the aspect in Eq. (\ref{n-expectation}). Theoretically, when $h$ takes on a large value, the model ends up with an enormous number of computing memory, making it challenging to achieve convergence. Conversely, when $h$ is excessively small, accurately extracting aspects becomes non-trivial.
Considering that aspect term lengths are typically not extensive, we conduct experiments by varying $h$ within the range of $\{1,2,3,4\}$. In our experimental results, presented in Figure~\ref{para2}, we showcase F1 scores for both the Restaurant and Laptop domains. As $h$ increases, we observe a gradual rise in performance up to a peak, followed by a decline. This suggests that maintaining a moderate $h$ value (such as 2 or 3) can not only reduce the parameter count but also ensure accurate aspect extraction.

\noindent
\textbf{Effect of parameter $\lambda$}.
We also analyze the effect of the contrastive weight $\lambda$ on F1-score across two datasets: Restaurant and Laptop. The results indicate that the optimal performance is achieved when $\lambda = 0.1$. As $\lambda$ increases beyond this point, the F1-score decreases steadily for both datasets, with a sharper decline in the Restaurant domain. This suggests that a smaller $\lambda$ effectively balances the contrastive and task-specific objectives, while higher values overly emphasize the contrastive loss, negatively impacting overall performance.

\subsection{Case Study and Error Analysis (RQ5)}
\textbf{Case Study}.  To offer a deeper understanding of FCKT's exceptional performance, we present several case studies in a qualitative manner in Figure~\ref{case}.
In 1st  case, the aspect detection and sentiment prediction in AIFI may be compromised due to weak sequential associations. For instance, it successfully identifies the aspect ``chefs” but falters in accurately predicting the sentiment associated with ``chefs” in the 1st case. Conversely, FCKT allows for the backpropagation of sentiment information into the boundary detection process, thereby bolstering correlations and leading to accurate  predictions.
By modeling semantic compatibility for aspect extraction via token-level contrastive learning, FCKT can extract more precise aspects. For example, in the 2nd case, word  ``separate”  should not be included in the aspect of ``ram memory”, a distinction that FCKT successfully identifies. 

\textbf{Error Analysis}. While the FCKT model has shown favorable performance, it still encounters challenges when handling complex sentences. For example, in the 3th sentence, due to the ambiguously expressed sentiment associated with ``restaurant”, the model struggles to make precise predictions.

\section{Conclusion}
In this paper, we addressed the problem of targeted sentiment analysis (TSA) by proposing a fine-grained cross-task knowledge transfer (FCKT) framework. By explicitly integrating aspect-level information into sentiment prediction, FCKT enables fine-grained knowledge transfer, effectively mitigating negative transfer and improving task performance. Extensive experiments conducted on three real-world benchmark datasets, including comparisons with diverse baselines and state-of-the-art LLMs validate the effectiveness of FCKT.

\section*{Acknowledgments}
We sincerely thank all the anonymous reviewers for their valuable comments to improve this paper. 
This work was supported by the National Key Research and Development Program of China under Grant
Nos. 2024YFF0729003, the National Natural Science Foundation of China under Grant Nos. 62176014, 62276015,
62206266, the Fundamental Research Funds for the Central Universities.
\section*{\centering Appendix}
\section*{A. Spliting Strategy}
In our modeling process, sentences containing multiple aspect terms are split into separate sentences during training, with each focusing on a single aspect, as illustrated in Figure \ref{split}. Here, we demonstrate that this strategy maintains optimization consistency while enabling fine-grained modeling of aspect-specific interactions.

First, when there are multiple aspects within a sentence, The original optimization objective $\mathcal{L}$ can be rewritten as:
\begin{equation}
	\begin{aligned}
		\mathcal{L}_{mul}&=\mathcal{L}_{a e}+\mathcal{L}_{cl}+\mathcal{L}_{sp} \quad \\
		&=-\sum_{i=1}^N \sum_{j=1}^m\left\{\boldsymbol{p}_{s, i, j}^T \log \left(\hat{\boldsymbol{p}}_{s, i, j}\right)+\boldsymbol{p}_{e, i, j}^T \log \left(\hat{\boldsymbol{p}}_{e, i, j}\right)\right\}\\
		&~~~~-
		\sum_{i=1}^N \sum_{j=1}^l\{\bm{s}(\bm{h}_s, \bm{h}_e)+\sum_{(x, y) \in O} \bm{s}(\bm{h}_x, \bm{h}_y)\}\\
		&~~~~-\sum_{i=1}^N \sum_{j=1}^l \sum_{k=1}^t \bm{y}_{i, j, k} \log \hat{\bm{y}}_{i, j, k} \quad\\
		&=- \sum_{i=1}^N \sum_{j=1}^m \left( \textcolor{white}{\Bigg|} \!\!\!\!\left\{\boldsymbol{p}_{s, i, j}^T \log \left(\hat{\boldsymbol{p}}_{s, i, j}\right)+\boldsymbol{p}_{e, i, j}^T \log \left(\hat{\boldsymbol{p}}_{e, i, j}\right)\right\}\right.\\
		&~~~~+\{\bm{s}(\bm{h}_s, \bm{h}_e)-\sum_{(x, y) \in O} \bm{s}(\bm{h}_x, \bm{h}_y)\}\\
		&\left.~~~+ \sum_{k=1}^t \bm{y}_{i, j, k} \log \hat{\bm{y}}_{i, j, k}\right),
	\end{aligned}
\end{equation}
where $N$ is the number of sentences, $m$ denotes the sentence's length, and $l$ is the number of aspects in each sentence. $\bm{s}(\cdot, \cdot)$ represents the cosine similarity function used in contrastive learning, defined as $\bm{s}(\bm{h}_x, \bm{h}_y) = \frac{\bm{h}_x \cdot \bm{h}_y}{\|\bm{h}_x\| \|\bm{h}_y\|}$, where $\bm{h}_x$ and $\bm{h}_y$ are embeddings of the tokens. $O$ denotes the set of negative pairs for each aspect, containing pairs of tokens that are not semantically aligned. $t$ refers to the number of sentiment categories (e.g., positive, neutral, and negative). Finally, $\bm{y}_{i, j, k}$ is the ground-truth label for the $k$-th sentiment category of the $j$-th aspect in the $i$-th sentence, and $\hat{\bm{y}}_{i, j, k}$ is the predicted probability distribution for the same. 

Building on the derivation above, it becomes evident that span-based targeted sentiment analysis can be interpreted as optimizing for a single aspect at a time. This formulation ensures that the optimization process remains unaffected by the number of aspects present in a sentence, allowing for consistent and efficient learning. Furthermore, under the assumption that each sentence contains only a single aspect, the optimization objective can be redefined as:
\begin{equation}
	\begin{aligned}
		\mathcal{L}_{sep}&=\mathcal{L}_{a e}+\mathcal{L}_{cl}+\mathcal{L}_{sp} \quad \\
		&=- \sum_{i=1}^M \sum_{j=1}^m \left( \textcolor{white}{\Bigg|} \!\!\!\!\left\{\boldsymbol{p}_{s, i, j}^T \log \left(\hat{\boldsymbol{p}}_{s, i, j}\right)+\boldsymbol{p}_{e, i, j}^T \log \left(\hat{\boldsymbol{p}}_{e, i, j}\right)\right\}\right.\\
		&~~~~+\{\bm{s}(\bm{h}_s, \bm{h}_e)-\sum_{(x, y) \in O} \bm{s}(\bm{h}_x, \bm{h}_y)\}\\
		&\left.~~~+ \sum_{k=1}^t \bm{y}_{i, j, k} \log \hat{\bm{y}}_{i, j, k}\right),
	\end{aligned}
\end{equation}
where $M$ represents the total number of aspects in the original training dataset, defined as $M = \sum_{i=1}^{N} l_{i}$, where $l_{i}$ denotes the number of aspects in the $i$-th sentence. The relationship between the two optimization objectives $	\mathcal{L}_{mul}$ and $	\mathcal{L}_{sep}$ can be expressed as:
\begin{equation}
	\mathcal{L}_{mul} \propto \mathcal{L}_{sep}.
\end{equation}

This indicates a direct proportionality between the two objectives, validating the effectiveness of our partitioning strategy. By leveraging this relationship, our method simplifies the training process while enhancing the model's capacity to manage aspect-specific interactions effectively, demonstrating its utility in fine-grained sentiment analysis.

\section*{B. Datasets}
To fairly assess our proposed FCKT, we conducted our experiments using three public datasets, Laptop, Restaurant, and Tweets. (1) The first dataset, referred to as Laptop, consists of customer reviews in the electronic product domain. It was collected from the SemEval Challenge 2014~\cite{pontiki2014semeval}. (2) The second dataset, named Restaurant, comprises reviews from the restaurant domain. It is a combination of review sets from SemEval2014, SemEval2015, and SemEval2016~\cite{pontiki2014semeval,pontiki2015semeval,pontiki2016semeval}.  (3) The third dataset, called Tweets, was created by Mitchell et al.~\cite{mitchell2013open}. It comprises Twitter posts from various users.	

\section*{C. Baselines}
In this study, we compare three categories of models:

\noindent
(1) \textbf{\underline{Pipeline-based models}}, which process tasks in sequential steps. These include:  
\begin{itemize}[leftmargin=*,labelindent=0pt,topsep=2pt,itemsep=1pt] 
	\item  CRF-Pipeline~\cite{mitchell2013open}: This paradigm employs a CRF as an aspect sequence extractor, succeeded by a sentiment classifier to achieve the objective.
	\item  NN-CRF-Pipeline~\cite{zhang2015neural}: Unlike the aforementioned model, this paradigm incorporates a shallow neural network model preceding the CRF.
	\item TAG-Pipeline~\cite{hu2019open}: It is a sequence tagging approach utilizing a BERT encoder.
	\item  SPAN-Pipeline~\cite{hu2019open}. It utilizes BERT as the shared encoder for two tasks and subsequently builds its own model for each task.
\end{itemize}

\noindent
(2) \textbf{\underline{End-to-End models}}, which directly map inputs to outputs without intermediate steps. Representative models are:  

\begin{itemize}[leftmargin=*,labelindent=0pt,topsep=2pt,itemsep=1pt] 
	\item  SPJM~\cite{zhou2019span}: It is a span-based method, which directly heuristically searches the boundaries of the aspect terms and then classifies the extracted aspect boundaries.
	\item  SPAN-Joint~\cite{hu2019open}: It utilizes BERT as the shared encoder for two tasks and subsequently builds its own model for each task.
	\item S-AESC~\cite{lv2021span}: The aspects and sentiments are generated collaboratively using both dual gated recurrent units and
	an interaction layer.
	\item  HI-ASA~\cite{chen2024modeling}: A hierarchical interactive network is devised with the aim of enhancing the mutual interactions
	between aspect and sentiment. This network incorporates input-side interactions as well as output-side interactions, forming a two-way communication framework.
	\item  DCS~\cite{li2023dual}: It proposes a dual-channel span generation method to effectively constrain the search space for span candidates for aspect-sentiment triplet extraction.
	\item  MiniConGTS~\cite{sun-etal-2024-minicongts}: This work proposes a minimalist tagging scheme and token-level contrastive learning strategy to improve pretrained representations for aspect sentiment triplet extraction, achieving promissing performance with reduced computational overhead.
	\item  PDGN~\cite{zhu2024pinpointing}: It introduces the grid noise diffusion pinpoint network , a T5-based generative model with three novel modules to address generation instability and improve robustness and effectiveness in aspect sentiment quad prediction tasks.
	\item AIFI~\cite{chen2024modeling}: The current state-of-the-art model for targeted sentiment analysis. AIFI is a variant of HI-ASA, introducing an adaptive feature interaction framework that leverages contrastive learning.
\end{itemize}

\noindent
(3) \textbf{\underline{LLM-based models}}, which leverage large language models with advanced reasoning capabilities. We leverage two large language model backbones, GPT-3.5-turbo and GPT-4o, employing four distinct techniques: 
\begin{itemize}[leftmargin=*,labelindent=0pt,topsep=2pt,itemsep=1pt] 
	\item  Zero-Shot: The model predicts without any prior task-specific examples, relying solely on its pre-trained knowledge to understand and perform the task.
	\item  Few-Shot: The model is provided with a small number of task-specific examples to guide its predictions, improving accuracy over zero-shot scenarios. We utilized 5-shot, 10-shot, and 20-shot methods, all randomly sampled from the training set. The results indicate that the 5-shot method performed the best, while the performances of the 10-shot and 20-shot methods showed a decline. The tables presents the output results for the 5-shot method.
	\item CoT: This approach involves generating intermediate reasoning steps, enabling the model to perform complex tasks by breaking them into smaller, logical steps.
	\item  CoT + Few-Shot: Combines the strengths of Chain of Thought reasoning and few-shot learning by providing task-specific examples along with explicit reasoning steps to enhance performance further.
\end{itemize}

\section*{D. Evaluation}
We employ three widely used metrics—precision, recall, and F1 score—to evaluate the effectiveness of our proposed FCKT. For aspect extraction, we focus on the F1 score as the primary evaluation metric, while accuracy is adopted for sentiment prediction. Notably, a predicted target is considered correct only if it exactly matches both the gold-standard aspect and its corresponding polarity.
To ensure robust and consistent evaluation while minimizing the impact of randomness, we conducted 10 independent experiments and report the average performance across these runs.

\section*{E. Implementation Details}
Following previous works \cite{lv2023efficient,chen2024modeling}, we split the training and test sets for each dataset. For the Tweets dataset, which lacks a predefined train-test split, we perform ten-fold cross-validation to ensure robust evaluation.
In the proposed model, we utilize the BERT-Large model as the backbone network. The maximum length $h$ of an aspect is selected from the range $\{1, 2, 3, 4\}$. The ratio $\xi$ of distributional input samples is varied between 0 and 1. Model optimization is conducted using the Adam optimizer~\cite{kingma2014adam}, with the learning rate searched from $\{2\text{e-}5, 2\text{e-}3, 2\text{e-}1\}$. The batch size is set to 16, and a dropout probability of 0.1 is applied. Additionally, the weight $\lambda$ is tuned over the range $\{0.1, 0.3, 0.5, 0.7\}$.
All experiments are implemented using the PyTorch framework and conducted on Nvidia GeForce Titan RTX 3090 GPUs, ensuring efficient training and reliable performance.

\bibliographystyle{named}
\bibliography{ijcai25}

\end{document}